\documentclass[review]{elsarticle}

\usepackage{lineno}
\modulolinenumbers[1]

\usepackage{hyperref}

\hypersetup{
    colorlinks=true,
    linkcolor=blue,
    filecolor=magenta,      
    urlcolor=cyan,
    pdftitle={Overleaf Example},
    pdfpagemode=FullScreen,
    }
    
\usepackage{url}

\usepackage{amsmath}
\usepackage{amsthm}
\usepackage{array}
\usepackage{bm}
\usepackage{float}
\usepackage{graphicx}
\usepackage{booktabs}
\usepackage{lipsum}
\usepackage{multirow}
\usepackage[numbers]{natbib}
\usepackage{subfig}

\usepackage{amsmath,amsfonts,bm}

\newcommand{\sota}{\emph{state-of-the-art}}
\newcommand{\etal}{\textit{et al.}}

\newcommand{\lrVert}[1]{\lVert #1 \rVert}

\theoremstyle{plain}

\theoremstyle{definition}

\theoremstyle{remark}

\newcommand{\reffig}[1]{Fig.~\ref{#1}}
\newcommand{\reftab}[1]{Tab.~\ref{#1}}

\def\eqref#1{equation~\ref{#1}}

\def\1{\bm{1}}

\def\rve{{\mathbf{e}}}

\def\rvg{{\mathbf{g}}}
\def\rvh{{\mathbf{h}}}

\def\rvm{{\mathbf{m}}}

\def\rvv{{\mathbf{v}}}

\def\rvx{{\mathbf{x}}}
\def\rvy{{\mathbf{y}}}

\def\bfM{{\mathbf{M}}}

\def\bfR{{\mathbf{R}}}

\def\bfU{{\mathbf{U}}}

\def\bfX{{\mathbf{X}}}
\def\bfY{{\mathbf{Y}}}

\DeclareMathOperator*{\argmin}{arg\,min}

\newcommand{\rom}[1]{%
  {\uppercase\expandafter{\romannumeral#1}}%
}

\journal{Journal of \LaTeX\ Templates}

\usepackage{numcompress}\biboptions{authoryear}

\begin{document}

\begin{frontmatter}

\title{Graph Neural Networks-based Hybrid Framework For Predicting Particle Crushing Strength}

\author{Tongya Zheng\fnref{zucc,cszju}}
\ead{doujiang_zheng@163.com}

\author{Tianli Zhang\fnref{softzju}}
\ead{zhangtianli@zju.edu.cn}

\author{Qingzheng Guan\fnref{cezju}}
\ead{11912051@zju.edu.cn}

\author{Wenjie Huang\fnref{cszju}}
\ead{wjie@zju.edu.cn}

\author{Zunlei Feng\fnref{softzju}}
\ead{zunleifeng@zju.edu.cn}

\author{Mingli Song\fnref{cszju}\corref{corres}}
\cortext[corres]{Corresponding author}
\ead{brooksong@zju.edu.cn}

\author{Chun Chen\fnref{cszju}}
\ead{chenc@zju.edu.cn}

\fntext[zucc]{Big Graph Center, College of Computer Science, Hangzhou City University}
\fntext[cszju]{College of Computer Science, Zhejiang University}
\fntext[softzju]{College of Software Technology, Zhejiang University}
\fntext[cezju]{Department of Civil Engineering, Zhejiang University}





\begin{abstract}
    Graph Neural Networks have emerged as an effective machine learning tool for multi-disciplinary tasks such as pharmaceutical molecule classification and chemical reaction prediction, because they can model non-euclidean relationships between different entities.
    Particle crushing, as a significant field of civil engineering, describes the breakage of granular materials caused by the breakage of particle fragment bonds under the modeling of numerical simulations, which motivates us to characterize the mechanical behaviors of particle crushing through the connectivity of particle fragments with Graph Neural Networks (GNNs).
    However, there lacks an open-source large-scale particle crushing dataset for research due to the expensive costs of laboratory tests or numerical simulations.
    Therefore, we firstly generate a dataset with 45,000 numerical simulations and 900 particle types to facilitate the research progress of machine learning for particle crushing.
    Secondly, we devise a hybrid framework based on GNNs to predict particle crushing strength in a particle fragment view with the advances of \sota~GNNs.
    Finally, we compare our hybrid framework against traditional machine learning methods and the plain MLP to verify its effectiveness.
    The usefulness of different features is further discussed through the gradient attribution explanation \emph{w.r.t} the predictions.
    Our data and code are released at \url{https://github.com/doujiang-zheng/GNN-For-Particle-Crushing}.
\end{abstract}


\end{frontmatter}

\section{Introduction}

Graph, as an abstract data structure, describes the complex relationships among various entities, such as the billions of directed links between webpages on the Internet~\cite{page1999pagerank}, the chemical bonds between atoms in the chemical engineering~\cite{gilmer2017neural}, the COVID-19 spread between people in the social networks~\cite{chang2021mobility}.
Recently, with the development of deep learning techniques~\cite{lecun2015deep}, Graph Neural Networks (GNNs)~\cite{kipf2016semi,bronstein2017geometric} have advanced the representation learning of graphs in multiple disciplinary tasks~\cite{gilmer2017neural,42vlassis2020geometric,zhang2022protein} by stacking non-linear graph convolution layers over the graph topology.
In civil engineering, numerical simulations usually treat a particle as a combination of several randomly tessellated particle fragments~\cite{hu2022particle,14de2014triaxial,15fu2017discrete,13cheng2004crushing,26huillca2021modelling}, and investigate the mechanical behaviors of particles under the external load based on the breakage of fragment bonds, which motivates us to handle the particle prediction tasks with the \sota~GNNs.

In the civil engineering, particles would break into several smaller fragments under the structural loads due to the surface flaws~\cite{ma2022morphology} or the internal flaws~\cite{16ma2014modeling,26huillca2021modelling,30griffith1921vi}, leading to the degradation of mechanical properties~\cite{4lade1996significance,5nakata1999probabilistic,6ovalle2014effect,cantor2017three}, which could cause serious consequences such as the excessive and dangerous settlement of buildings~\cite{1hardin1985crushing,2mcdowell1998micromechanics,3zhu2019modeling,4lade1996significance,5nakata1999probabilistic,6ovalle2014effect,24nader2019effect}.
Recent research of particle crushing has largely extended the investigation of particle crushing from a single particle~\cite{mcdowell2000application} to particles with different mechanical properties such as particle sizes~\cite{10huang2014size,26huillca2021modelling}, particle shapes~\cite{15fu2017discrete,hu2022particle,zhu2021interplays}, and particle morphology~\cite{sun2021influence,22wang2021machine,ma2022morphology}.
These dedicated heuristics of particle properties are summarized as morphology descriptors of particles in~\cite{22wang2021machine}, which can be divided into the form descriptors, the roundness descriptors, and the sphericity descriptors.

Despite the effectiveness of morphology descriptors, they are far away from the particle crushing mechanisms investigated by existing numerical simulation methods~\cite{hu2022particle,14de2014triaxial,15fu2017discrete,13cheng2004crushing,26huillca2021modelling}.
These methods model a particle as a polyhedron of the connected fragments and obtain the particle states at each time step by solving the displacements of different fragments, which investigates particle crushing in a microscopic view instead of the morphology view.
The simulated particle crushing process inspires us to model the particle as a connected fragment graph and advance the representation of particles from morphology descriptors to their internal mechanisms.
Furthermore, unlike the machine learning community that has released various datasets, there lacks an open-source dataset of particle crushing, which significantly hinders the development of its machine learning research.

Therefore, we firstly provide an open-source large-scale particle crushing dataset of 45,000 particles by simulating the crushing process of a single rockfill particle under one-dimensional (1D) compression with the Non-Smooth Contact Dynamics (NSCD) method following~\cite{hu2022particle}.
The particle crushing dataset consists of 900 different particle types, with 20 different particle diameters, 15 different particle shapes, and 3 different compression axes.
Each type of particle is randomly tessellated into multiple polyhedron fragments by the Neper~\cite{27kun1996study,28nguyen2015bonded,29quey2011large} with the specific diameter and shape.
The tessellated particle is then compressed by two rigid parallel platens along the specific compression axis, following a DEM model with a three-dimensional bonded-cell method~\cite{26huillca2021modelling}.
The numerical model was implemented in the software LMGC90~\cite{dubois2013lmgc90}, which is based on the non-smooth contact dynamics method~\cite{32rafiee2008modelling,33rafiee2011stochastic}. 

Secondly, we model the particle fragments as a connected fragment graph to utilize \sota~machine learning tools for particle representation.
The interacted fragments of a particle could be seen as a connected graph, where each fragment acts as a vertice and interacts with other fragments during the particle crushing process. 
The complex interaction between fragments could be learned by the emerging Graph Neural Networks (GNNs)~\cite{bronstein2017geometric,kipf2016semi,xu2018how} with the large-scale particle crushing dataset.
We are inspired by these encouraging applications of GNNs~\cite{feng2019meshnet,chang2020learning,schulte2021integration,ying2021transformers,yang2020breaking,zhang2021litegem,deepmind2021ogb,42vlassis2020geometric} to build a GNNs-based hybrid framework to combine GNNs based on the fragment graph with MLP based on the dedicated morphology descriptors, predicting the characteristic crushing strength of particles on the newly given particles in the testing set.

Finally, we have conducted experiments to examine the prediction performance of our proposed hybrid framework, the effectiveness of different components, and the attributions of the calculated features.
To validate the generalization ability of different methods, we propose three kinds of testing sets to construct three different tasks, which are split by the diameter, shape, and compression axis, respectively.
We have calculated geometric descriptors as described in~\cite{10huang2014size,26huillca2021modelling,15fu2017discrete,hu2022particle,zhu2021interplays,sun2021influence} and compared five traditional machine learning methods~\cite{sklearn,xgboost,lightgbm} (denoted as non-Deep methods) against a bunch of Deep Neural Networks (DNNs) including three kinds of \emph{state-of-the-art} GNNs~\cite{feng2019meshnet,xu2018how,yang2020breaking}, where GNNs are implemented using our hybrid framework.
The contributions of the morphology descriptors for MLP and the node and edge features for GNNs are further verified through the ablation studies on three GNNs across three different tasks.
The common and individual patterns of the three tasks are analyzed through the visualization results of feature attributions based on the gradient norms.

The main contributions can be summarized as follows:
\begin{itemize}
    \item We generate a large-scale particle crushing dataset to facilitate the machine learning research of this field, comprising 45,000 numerical simulations with 900 different particle types.
    \item We build a GNNs-based hybrid framework to model a particle with elementary cell interactions and its overall morphology descriptors for predicting the particle crushing strength across three different tasks.
    \item We have conducted elaborate comparison experiments and ablation studies to investigate the effectiveness of GNNs, particle morphology descriptors, and features for nodes and edges. We further visualize the common and individual patterns through the saliency map of different features.
\end{itemize}

\section{Related Works}

\subsection{Graph Neural Networks}

Graph Neural Networks (GNNs) have shown their powerful prediction abilities on the non-grid data and have been applied to various fields such as Recommendation~\cite{ying2018graph}, traffic flow prediction~\cite{yu2018spatio}, and molecular classification~\cite{gilmer2017neural,xu2018how,yang2020breaking,ying2021transformers}.
A graph is defined by a node set $V=\{v_1, \cdots, v_n\}$ and an associated edge set $E=\{e_1, \cdots, e_m\}$, where $e_1$ connects a pair of adjacent fragments $v_i$ and $v_j$.
The message-passing paradigm~\cite{gilmer2017neural} describes GNNs in a microscopic view that each node $v_i$ receives the messages from its neighbors $v_j \in \mathcal{N}(v_i)$ and updates its representation based on these messages, written as:
\begin{equation}
    \begin{split}
       \rvm_i^k &= \sum_{v_j \in \mathcal{N}(v_i)} \bfM_k(\rvh_i^k, \rvh_j^k, \rve_{ij}), \\
       \rvh_i^{k+1} &= \bfU_k(h_i^k, \rvm_i^k),
    \end{split}
\end{equation}
where $v_i$ updates from $\rvh_i^{k}$ of the $k$-th layer to $\rvh_i^{k+1}$ of $k+1$-th layer based on the message function $\bfM_k$ and the update function $\bfU_k$.
The obtained node representation $\rvh_i$ can be used for various tasks such as node classification~\cite{kipf2016semi} and link prediction~\cite{ying2018graph}.
It requires a readout function to generate a representation vector for the whole graph $G$ for graph-level tasks like molecular classification~\cite{gilmer2017neural,xu2018how,yang2020breaking,ying2021transformers}, written as
\begin{equation}
    \rvh_G = \bfR(\{\rvh_i^K \vert v_i \in V\}),
\end{equation}
where $K$ is the number of neural network layers. 
Furthermore, GNNs have proved their applications in various fields, such as predicting the deformation of 3D objects given the 3D meshed data points~\cite{feng2019meshnet}, generating artificial building structures to alleviate the efforts of experts~\cite{chang2020learning}, predicting the HOMO-LUMO energy gap of molecules given their 2D molecular graphs~\cite{schulte2021integration,ying2021transformers,yang2020breaking,zhang2021litegem,deepmind2021ogb}, and predicting the homogenized responses of polycrystals~\cite{42vlassis2020geometric}.


\subsection{Particle Crushing}

McDowell~\etal~\cite{mcdowell2000application} have found that the particle crushing strength of geometrically similar particles follows a Weibull distribution, written as
\begin{equation}
    P_s = \exp \left[ - \left(\frac{d_s}{d_0}\right)^3 \left(\frac{\sigma}{\sigma_0}\right)^m \right],
    \label{eq:weibull}
\end{equation}
where $P_s$ is the breakage ratios under the corresponding load, $d_0$ and $\sigma_0$ are the characteristic diameter and crushing strength, respectively.
The survival probability of a batch of 50 particles with the same geometry is drawn in \reffig{fig:sample-weibull}, sorting by the normalized particle strength, where $m=9.690$ is the Weibull modulus and indicates the variability of the particle crushing distribution.
The characteristic particle strength is thus scaled to 1.0 of the horizontal axis and corresponds to the 37\% survival probability of the vertical axis.
Existing particle crushing theories~\cite{mcdowell2000application,sun2021influence,22wang2021machine} derive a closed-form formula to describe the size effects of particles with different diameters.
However, firstly, their various theories have been proposed from different perspectives and lacked a consensus on particle crushing behaviors.
Secondly, our experimental results of the non-Deep methods and DNNs in \reftab{tab:main-result} reveal that particle morphology descriptors play a vital role in the generalization of predicting the particle crushing strength while existing theories are mostly based on the particle diameter.
Thirdly, generalizing the predictions for more conditions like particle shapes and compression axes requires a constitutive model to describe the particle from different viewpoints.
Therefore, we are motivated to model the fragment connection of the particle with Graph Neural Networks (GNNs), which offer natural modeling for the fragment graph.

\section{Method}

\begin{figure}[t]
    \centering
    \includegraphics[width=\textwidth]{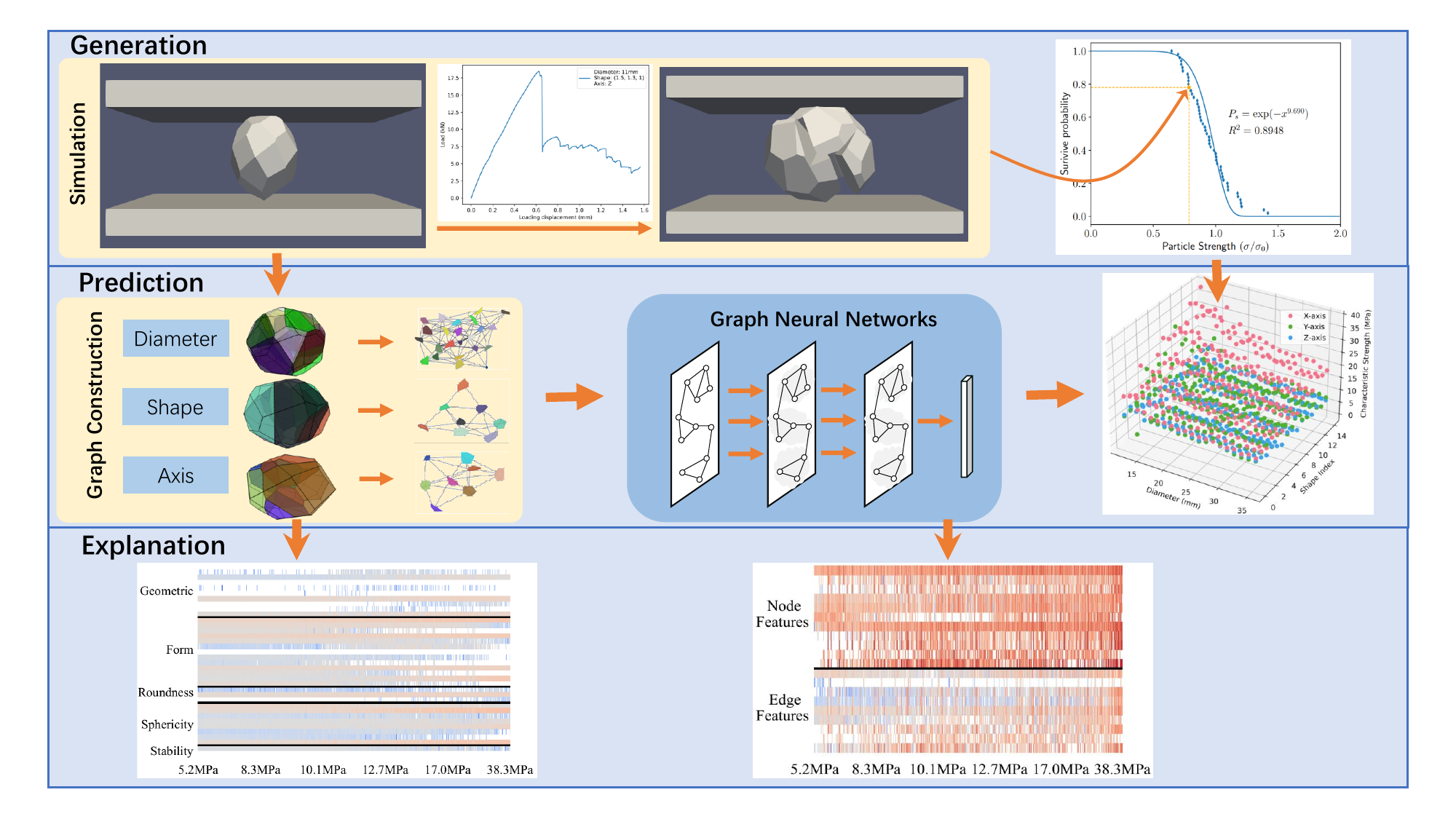}
    \caption{The pipeline of our work consists of three parts, where the \textbf{generation} part means the large-scale dataset, the \textbf{prediction} part means the graph construction of each particle and the hybrid framework for particle crushing strength, and the \textbf{explanation} part visualizes the importance of different features w.r.t their gradient magnitudes.}
    \label{fig:framework}
\end{figure}

As shown in~\reffig{fig:framework}, we first conduct numerical simulations for 900 particle types with different diameters, shapes, and compression axes to measure the characteristic particle crushing strength.
Secondly, we build a hybrid framework to combine the \sota~GNNs with the MLP network to predict the strength of the Diameter, Shape, and Axis tasks and validate the generalization ability of DNNs and GNNs.
Lastly, we quantitatively attribute the input contributions to the predictions and analyze the common and individual patterns of the three tasks.
\subsection{Dataset Generation}

\begin{table}[t]
    \caption{Configuration of particle crushing dataset. The \textit{italic} font refers to the testing settings.}
    \label{tab:dataset} 
    \resizebox{\linewidth}{!}{
        \centering
    \begin{tabular}{l|p{0.6\linewidth}}
        \hline
        Dimension & Setting \\
        \hline
        Diameter (mm) & 11.86, 13.10, 14.35, 15.60, 16.85, 18.10, 19.34,\
         
        20.59, 21.84, 23.09, 24.34, 25.58, 26.83, \textit{28.08},\
         
        \textit{29.33, 30.58, 31.82, 33.07, 34.32, 35.57} \\
        \hline
        Scale Shape (X, Y, Z) & (1,1,1),\textit{(1/0.95,0.95,1)},(1/0.9,0.9,1),\
        
        (1.1,1.1,1),\textit{(1.25,1.21/1.25,1)},(1.21/0.9,0.9,1),\
        
        (1.2,1.2,1),\textit{(1.25,1.44/1.25,1)},(1.5,1.44/1.5,1),\
        
        (1.3,1.3,1),\textit{(1.25,1.69/1.25,1)},(1.5,1.69/1.5,1),\
        
        (1.4,1.4,1),\textit{(1.25,1.96/1.25,1)},(1.5,1.96/1.5,1)  \\
        \hline
        Compression Axis & X-axis, \textit{Y-axis}, Z-axis \\ 
        \hline
    \end{tabular}
    }
\end{table}

To overcome the limitations of small-scale datasets of existing researches~\cite{10huang2014size,26huillca2021modelling,15fu2017discrete,hu2022particle,sun2021influence,18zhu2019peridynamic,22wang2021machine,28nguyen2015bonded}, we firstly generate 45,000 particles and simulate their crushing processes based on numerical simulations~\cite{hu2022particle,14de2014triaxial,15fu2017discrete,13cheng2004crushing,26huillca2021modelling}.
As shown in \reftab{tab:dataset}, there are 900 different particle types in total, the Cartesian product of 20 different particle diameters, 15 different scale shapes in the (X, Y, Z) axes, and 3 different compression axes under one-dimensional compression.
Each type contains 50 particle crushing tests~\cite{mcdowell2000application,hu2022particle,sun2021influence} to estimate the characteristic crushing strength of the specific particle type, where each test is based on a randomly tessellated particle.
The comprehensive large-scale dataset can measure the generalization ability of the {\sota} machine learning methods from the three aspects of diameter, shape, and axis, and examine the feature attributions with respect to the predictions of particle crushing strength based on the well-trained models.

The top row of \reffig{fig:framework} describes the crushing process of an example particle compressed by two plates, where the particle stress drops drastically from the peak when the particle breaks into pieces~\cite{mcdowell2000application,26huillca2021modelling} as shown in the stress curve with respect to the time step.
The particle strength is computed by $\sigma = F_c/d^2$, where $F_c$ is the peak stress and $d$ is the distance between two plates, serving as a data point in a batch of particle crushing tests of a specific particle type, which is given the specific diameter, shape, and axis.
As shown in the top right plot of \reffig{fig:framework}, the batch of particle crushing tests fits the Weibull statistics by sorting the particle strength in an ascending order, where the particle strength is normalized by the characteristic particle strength $\sigma_0$ at 37\% survival probability.
When increasing the particle strength, the survival probability $P_s$ of the batch of particles decreases exponentially fast at $P_s = \exp(-x^{9.690})$ with $R^2=0.8948$, where the Weibull modulus $9.690$ depicts the sharpness of the strength-probability curve.
As shown in previous researches~\cite{10huang2014size,26huillca2021modelling,15fu2017discrete,hu2022particle,sun2021influence,18zhu2019peridynamic,22wang2021machine,28nguyen2015bonded}, when individually and jointly varying the conditions of diameter, shape, and axis, different particle types exhibit distinct mechanical properties.
Therefore, it is of vital importance to generalize prediction models from the comprehensive training set to the unseen testing set, capturing the macroscopic properties of granular materials.

In the numerical simulation of a particle compression test, a randomly tessellated particle by Neper~\cite{27kun1996study,28nguyen2015bonded,29quey2011large} is firstly meshed by Gmsh~\cite{gmsh} and then compressed by two rigid plates using LMGC90~\cite{dubois2013lmgc90} with a Non-Smooth Contact Dynamics (NSCD)~\cite{31jean1999non,32rafiee2008modelling,33rafiee2011stochastic} method. 
Firstly, a batch of 50 particles is randomly tessellated into fragments for a specific particle type following the Voronoi tessellation algorithm~\cite{29quey2011large} to simulate the fragment size distributions of realistic rockfill particles.
Secondly, the numerical model implemented by LMGC90~\cite{dubois2013lmgc90} solves the displacements of particle fragments by computing the contact forces of each body following the Signorini-Coulomb condition~\cite{jean1992unilaterality,moreau1988unilateral} and the Cohesive Zone Model (CZM) model~\cite{hu2022particle,26huillca2021modelling}.
The simulated particle breaks into pieces when the CZM bonds of particle fragments break due to external loadings.
As shown in \reffig{fig:framework}, the peak force of particle crushing is recorded when a drastic drop of more than 35\% of the force is observed in the time step-force curve, indicating the particle breakage phenomenon~\cite{22wang2021machine,26huillca2021modelling,hu2022particle}. 

Finally, we summarize the parameter settings of the particle generation process as follows.
In the particle tessellation process, the number of particle fragments is initialized by 8 with a diameter of 11.86 mm and increases cubically from 8 to 216 with respect to the diameter, ensuring the constant cell density following that the internal flaws of particles increase with the increasing particle size~\cite{30griffith1921vi}.
In the particle crushing process, the friction coefficient between the rigid plates and the particle is set to 0.3 with a constant vertical velocity of the upper plate, the time step is set to $5\times 10^{-4}$s, and the CZM parameters of particles are calibrated by the commonly adopted Brazilian splitting test~\cite{jiang20183d,yu2004energy}.
To examine the generalization ability of machine learning methods, the ratios of the testing set are set to 7/20, 5/15, 1/3 in terms of different diameters, shapes, and axes, respectively.

\subsection{Problem Setup of Predicting Particle Crushing Strength}

Given a detailed 3D mesh of a particle and its fragments, the regression task of predicting the characteristic particle crushing strength $\sigma_0$ is written as
\begin{equation}
    \argmin_{\mathcal{F}}  \sum_{\rvx \in \bfX, \rvy \in \bfY} \lrVert{\rvy - \mathcal{F}(\rvx)}_2 + \lambda\Omega(\mathcal{F}),
    \label{eq:reg-loss}
\end{equation}
where $\rvx, \rvy$ are the particle and its $\sigma_0$, $\bfX, \bfY$ are the sets of input particles and their prediction labels, $\mathcal{F}$ is a machine learning method, the task is optimized by the Mean Squared Error (MSE) loss between the labels $\rvy$ and the predictions $\mathcal{F}(x)$ and a regularization term $\Omega(\mathcal{F})$ reducing the complexity of $\mathcal{F}$, and $\lambda$ is a tunable coefficient. 

Based on the 3D mesh $\rvx$ of a particle, we follow the particle morphology descriptors of existing researches~\cite{blott2008particle,bagheri2015characterization,huang2020experimental,domokos2015universality,pouranian2020impact,zhao20163d,yang2017effects,zheng2021three,zhang2016preliminary}, which have shown dominant impacts on the crushing strength.
As summarized by Wang~\etal~\cite{22wang2021machine}, there are thirty-five particle morphology descriptors, including nine geometric characteristics~\cite{22wang2021machine}, thirteen form descritpors~\cite{blott2008particle,bagheri2015characterization,huang2020experimental,domokos2015universality}, three roundness descritpors~\cite{pouranian2020impact,huang2020experimental,zhao20163d}, nine sphericity descriptors~\cite{blott2008particle,pouranian2020impact,yang2017effects,zheng2021three,huang2020experimental,zhang2016preliminary}, and one instability descriptor~\cite{22wang2021machine}. 
These particle morphology descriptors are fed into traditional machine learning methods for predicting the particle crushing strength, including Linear Regression, Ridge Regression, Random Forest~(RF), XGBoost~(XGB)~\cite{xgboost}, and LightGBM~(LGB)~\cite{lightgbm}.
The former two are linear methods, and the latter three are ensemble tree-based methods, which can capture complex patterns of morphology descriptors.
We further apply a Multi-Layer Perceptron~(MLP) to model the multivariate relationships of various morphology descriptors, which serves as the baseline method to verify the effectiveness of deep learning methods and Graph Neural Networks~(GNNs). 

\subsection{GNNs-based Hybrid Framework}

Despite their significant efforts, morphology descriptors mostly work based on insightful observations of researchers and hardly take the internal structure of particles into account, which plays an essential role in the particle crushing process as shown in previous literatures~\cite{2mcdowell1998micromechanics,7daouadji2001elastoplastic,8russell2004bounding,9tengattini2016constitutive,ma2022morphology}.
In a numerical simulation, a particle is divided into multiple fragments randomly~\cite{29quey2011large} and breaks into pieces due to the breakage of the CZM bonds of fragments~\cite{hu2022particle}.
The mutual relationships of particle fragments thus influence the particle crushing behaviors significantly, which can doubtlessly help the predictions if the internal structure is considered.

Let $V=\{v_1, \cdots, v_n\}$ be the fragment set of a particle, where $v_1$ is one of the fragments, and $n$ is the number of fragments.
The mutual relationships of $V$ can be well described by an edge set $E=\{e_1, \cdots, e_m\}$, where $e_1$ connects a pair of adjacent fragments $v_i$ and $v_j$.
The consequent fragment graph $G=\{V, E\}$ describes the internal structure of a particle, which can output the latent representation of the particle using \sota~GNNs~\cite{feng2019meshnet,xu2018how,yang2020breaking}.
As shown in \reffig{fig:framework}, GNNs use the fragment graph of particles as the inputs and predict the corresponding particle crushing strength $\sigma_0$ in terms of different diameters, shapes, and axes, which can be written as
\begin{equation}
    \argmin_{\rvg}  \sum_{\rvx \in \bfX, \rvy \in \bfY} \lrVert{\rvy - \rvg(\rvx, G)}_2 + \lambda\Omega(\rvg),
\end{equation}
where $\rvg$ is one kind of GNNs.
On the one hand, GNNs can summarize the latent representations from the fragment level to the particle level by aggregating neighborhood information iteratively with the fragment graph, complementing the missing microscopic interactions of fragments of particle morphology descriptors.
On the other hand, dedicated efforts for particle morphology descriptors can largely alleviate the task difficulty of predicting particle crushing strength $\sigma_0$, which is verified by \reffig{fig:ablation}. 
Therefore, we build a hybrid framework of a GNN and an MLP by summing their predictions to advance the training of GNNs, written as $\rvg^{\prime}=\rvg(\rvx, G) + \mathrm{MLP}(\rvx)$.

We further prepare eleven node features and nine edge features for the fragment graph to characterize the internal states of particles, where a node is a particle fragment and an edge is the interaction between a pair of fragments.
Following Vlassis~\etal~\cite{42vlassis2020geometric}, the states of a node are represented by its volume, its external area, diameter, the number of its external faces, the number of its neighborhood fragments, the 3D coordinates of its centroid, and its three Euler angles, which are computed by its mesh file generated by Neper~\cite{29quey2011large}.
Similarly, the edge states are represented by the contact area of fragments, the number of lines of the contact area, the maximum length of the contact area, the 3D coordinates of its centroid, and its three Euler angles.
These node and edge features are then fed into GNNs for the latent representations of particles.

\section{Experiment}

\subsection{Performance Comparison}

\begin{table}
    \caption{The distribution of particle crushing strength in the testing sets.}
    \label{tab:test-dist}
    \centering
    \begin{tabular}{cccc}
        \toprule
        Quantile & \textbf{Diameter} (MPa) & \textbf{Shape} (MPa) & \textbf{Axis} (MPa) \\
        \hline
        min  &  5.09 &  5.54 &  5.09 \\
        25\% &  8.05 &  9.09 &  7.49 \\
        50\% & 10.13 & 11.27 &  8.80 \\
        75\% & 13.38 & 14.70 & 10.04 \\
        max  & 21.71 & 33.34 & 17.62 \\
        \bottomrule
    \end{tabular}
\end{table}

\begin{table*}
    \caption{Particle crushing prediction performance.}
    \label{tab:main-result}
    \centering
    \resizebox{\textwidth}{!}{
    \begin{tabular}{lcc|cc|cc}
        \toprule
        & \multicolumn{2}{c}{ \textbf{ Diameter }}  & \multicolumn{2}{c}{ \textbf{ Shape }}  & \multicolumn{2}{c}{ \textbf{ Axis }}\\ 
        \cline{2-7}
        & MAE (MPa) & RMSE (MPa) & MAE (MPa) & RMSE (MPa) & MAE (MPa) & RMSE (MPa) \\
        \hline
        Linear & $ 3.961 \pm 0.000 $  & $ 4.511 \pm 0.000 $  & $ 2.266 \pm 0.000 $  & $ 2.924 \pm 0.000 $  & $ 2.693 \pm 0.000 $  & $ 3.259 \pm 0.000 $ \\
        Ridge & $ 2.725 \pm 0.000 $  & $ 3.290 \pm 0.000 $  & $ 2.271 \pm 0.000 $  & $ 2.950 \pm 0.000 $  & $ 2.666 \pm 0.000 $  & $ 3.223 \pm 0.000 $ \\
        RF & $ 1.084 \pm 0.011 $  & $ 1.486 \pm 0.012 $  & $ 1.140 \pm 0.007 $  & $ 1.591 \pm 0.013 $  & $ 2.434 \pm 0.035 $  & $ 3.625 \pm 0.039 $ \\
        XGB & $ \bm{0.810 \pm 0.000} $  & $ \bm{1.084 \pm 0.000} $  & $ \bm{1.025 \pm 0.000} $  & $ 1.601 \pm 0.000 $  & $ \bm{1.451 \pm 0.000} $  & $ \bm{2.120 \pm 0.000} $ \\
        LGB & $ 0.881 \pm 0.000 $  & $ 1.207 \pm 0.000 $  & $ 1.047 \pm 0.000 $  & $ \bm{1.428 \pm 0.000} $  & $ 1.762 \pm 0.000 $  & $ 2.534 \pm 0.000 $ \\
        \hline
        MLP & $ 0.771 \pm 0.023 $  & $ 1.032 \pm 0.024 $  & $ 0.935 \pm 0.029 $  & $ \bm{1.316 \pm 0.056} $  & $ 0.814 \pm 0.079 $  & $ 1.197 \pm 0.101 $ \\
        MeshNet & $ 0.719 \pm 0.054 $  & $ 0.958 \pm 0.048 $  & $ 1.027 \pm 0.046 $  & $ 1.546 \pm 0.053 $  & $ 0.731 \pm 0.092 $  & $ 1.105 \pm 0.122 $ \\
        GIN & $ 0.719 \pm 0.083 $  & $ 0.951 \pm 0.070 $  & $ 0.910 \pm 0.051 $  & $ 1.368 \pm 0.095 $  & $ 0.746 \pm 0.065 $  & $ 1.204 \pm 0.054 $ \\
        ExpC & $\bm{ 0.709 \pm 0.025 }$  & $\bm{ 0.944 \pm 0.043 }$  & $ \bm{0.883 \pm 0.036} $  & $ 1.388 \pm 0.047 $  & $ \bm{0.728 \pm 0.031} $  & $ \bm{1.079 \pm 0.036} $ \\
        \bottomrule
    \end{tabular}
    }
\end{table*}

\begin{table}
    \caption{The p-values of ExpC and other DNNs under two-sided t-test.}
    \label{tab:pvalue}
    \centering
    \begin{tabular}{lccc}
        \toprule
        & \textbf{Diameter} & \textbf{Shape} & \textbf{Axis} \\
        \hline
        MLP     & 0.0035 & 0.0372 & 0.0531 \\
        MeshNet & 0.7191 & 0.0006 & 0.9443 \\
        GIN     & 0.8038 & 0.3681 & 0.5929 \\
        \bottomrule    
    \end{tabular}
\end{table}

We split different testing sets of different diameters, shapes, and axes to examine the task difficulty and generalization ability of methods.
The compared methods include the \sota~machine learning methods (denoted as non-Deep methods)~\cite{sklearn,xgboost,lightgbm} and the \sota~Deep Neural Networks (DNNs)~\cite{feng2019meshnet,xu2018how,yang2020breaking}, aims at validating whether DNNs are superior to non-Deep methods and whether GNNs are superior to other DNNs.
For three tasks of the Diameter, Shape, and Axis, we have tried numerous hyper-parameter settings of methods for a fair comparison of prediction ability and stability, where each setting takes 5 runs of fixed random seeds for reproducibility, as listed in \reftab{tab:nondeep-models}, \reftab{tab:default-deep-models}, and \reftab{tab:deep-models}. 
For non-Deep methods, there are 15 runs for Linear Regression, 60 runs for Ridge Regression, 675 runs for Random Forest, 5625 runs for XGBoost~\cite{xgboost}, and 5625 runs for LightGBM~\cite{lightgbm}.
For DNNs, there are 225 runs for MLP, 675 runs for MeshNet~\cite{feng2019meshnet}, 855 runs for GIN~\cite{xu2018how}, and 540 runs for ExpC~\cite{yang2020breaking}, where each setting takes 5 runs of fixed random seeds and 3 ablation studies of different inputs.
These detailed experiments can help us dive into the model generalization ability for predicting the particle crushing strength.

Overall, the generalization difficulty of both tasks and methods is examined that DNNs can overcome the difficulty of the Axis task by the non-linearity modeling, and the large prediction variances of GNNs cause its statistical insignificance among different GNNs.

\textit{The Axis task is the most difficult task for non-Deep methods, and the Shape task is the most difficult task for DNNs.}
Existing researches~\cite{10huang2014size,26huillca2021modelling,15fu2017discrete,hu2022particle,zhu2021interplays,sun2021influence} investigate the effects of the particle crushing strength based on either of the three tasks while neglecting the difficulty comparison between different tasks.
\reftab{tab:main-result} presents that non-Deep methods and DNNs both achieve the least Mean Absolute Error (MAE) and Root Mean Square Error (RMSE) on the Diameter task.
On the one hand, the size effect of particle crushing strength has been sufficiently investigated with morphology descriptors.
On the other hand, the larger size of particles leads to the degeneration of the strength due to the increasing internal flaws~\cite{10huang2014size,22wang2021machine}, as shown in~\reftab{tab:test-dist}.
However, although the maximum strength of the testing set of the Axis task is nearly half of that of the Shape task as shown in \reftab{tab:test-dist}, non-Deep methods present significantly larger MAE in the Axis task.
In contrast, MLP reduces the MAE by 43.9\% on the Axis task compared to XGBoost, the best non-Deep method, based solely on the morphology descriptors, indicating the effective non-linearity modeling of DNNs.
ExpC further outperforms MLP by 10.5\% with $p=0.0531$ at the two-sided independent t-test, verifying the usefulness of the fragment graph modeling.
The four methods of DNNs obtain higher MAE on the Shape task than that on the Axis task, which is consistent with the testing set distribution as shown in \reftab{tab:test-dist}.
It indicates that DNNs can better use input morphology descriptors and the fragment graph than non-Deep methods.

\textit{MLP outperforms non-Deep methods statistically more significantly.}
Since XGBoost and LightGBM are not affected by different random seeds, MLP achieves statistical significance than them on all three tasks and reduces the MAE by 4.8\% on the Diameter task, 8.8\% on the Shape task, and 43.9\% on the Axis task.
Furthermore, MLP shares the same morphology descriptors with non-Deep methods, indicating that the particle crushing theory with respect to different conditions requires relational modeling of different aspects, including the geometric, form, roundness, and sphericity descriptors.

\textit{ExpC outperforms MLP statistically more significant, but doesn't show statistical significance of other GNNs.}
\reftab{tab:pvalue} presents that ExpC achieves statistical significance of MLP with the help of the fragment graph and the powerful framework of GNNs.
However, although ExpC performs best among GNNs, the p-values of MAE of ExpC and that of other GNNs don't show statistical significance, which is caused by the large variances of other GNNs with different random seeds.
It implies a future direction to control the prediction variances of GNNs for predicting the particle crushing strength.


\subsection{Ablation Study}

\begin{figure}[htbp]
    \centering
    \includegraphics[width=\linewidth]{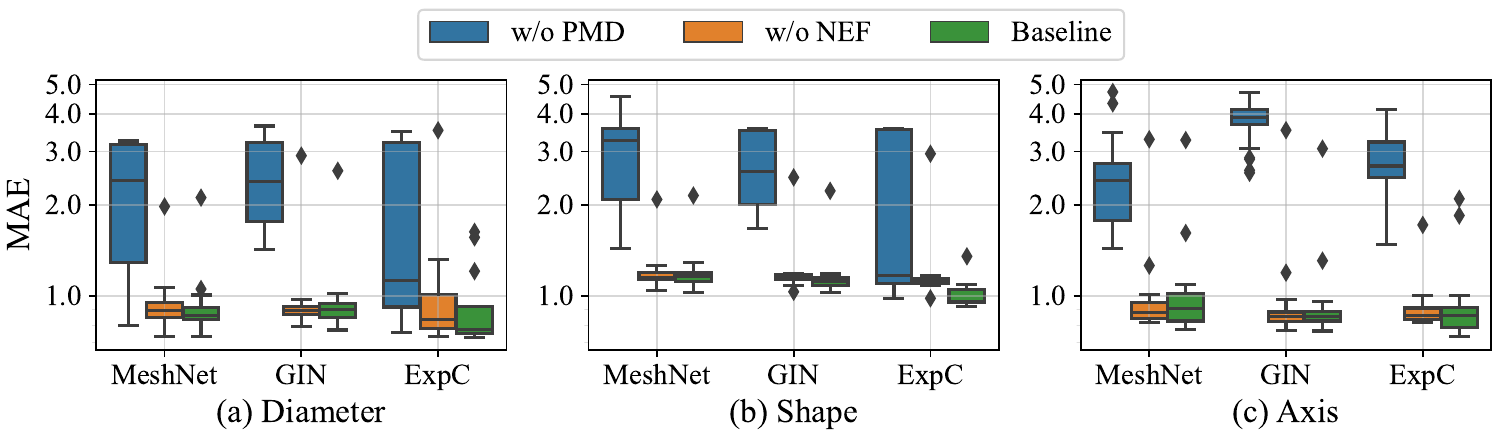}
    \caption{MAE distribution of ablation studies among all parameter settings, including GNNs w/o particle morphology descriptors (PMD), GNNs w/o node and edge features (NEF), and GNNs with both PMD and NEF (Baseline).}
    \label{fig:ablation}
\end{figure}

\begin{table}
    \caption{The best performance comparison of ablation studies among all parameter settings, including GNNs w/o particle morphology descriptors (PMD), GNNs w/o node and edge features (NEF), and GNNs with both PMD and NEF (Baseline).
    The degradation percentages compared to the Baseline are listed here.
    }
    \label{tab:ablation}
    \centering
    \resizebox{\linewidth}{!}{
    \begin{tabular}{cc|cc|cc|cc}
        \toprule
        Method & Ablation Study & Diameter & Deg. (\%) & Shape & Deg. (\%) & Axis & Deg. (\%) \\
        \hline
        \multirow{3}{*}{MeshNet} 
        & w/o PMD & $ 0.800 \pm 0.044 $ & -11.3 & $ 1.506 \pm 0.021 $ & -46.6 & $ 1.429 \pm 0.081 $ & -95.5 \\
        & w/o NEF & $ 0.733 \pm 0.073 $ & -1.9 & $ 1.038 \pm 0.108 $ & -1.1 & $ 0.815 \pm 0.053 $ & -11.5 \\
        & Baseline & $ 0.719 \pm 0.054 $  & - & $ 1.027 \pm 0.046 $ & - & $ 0.731 \pm 0.092 $ & - \\
        \hline
        \multirow{3}{*}{GIN} 
        & w/o PMD & $ 1.425 \pm 0.162 $  & -98.2 & $ 1.664 \pm 0.122 $  & -82.9 & $ 2.551 \pm 0.286 $ & -242.0 \\
        & w/o NEF & $ 0.789 \pm 0.028 $  & -9.7 & $ 1.030 \pm 0.058 $  & -13.2 & $ 0.768 \pm 0.043 $ & -2.9 \\
        & Baseline & $ 0.719 \pm 0.083 $  & - & $ 0.910 \pm 0.051 $  & - & $ 0.746 \pm 0.065 $ & - \\
        \hline
        \multirow{3}{*}{ExpC} 
        & w/o PMD & $ 0.759 \pm 0.072 $  & -7.1 & $ 0.982 \pm 0.039 $  & -11.2 & $ 1.477 \pm 0.228 $ & -102.9 \\
        & w/o NEF & $ 0.734 \pm 0.033 $  & -3.5  & $ 0.981 \pm 0.070 $  & -11.1 & $ 0.815 \pm 0.045 $ & -12.0 \\
        & Baseline & $ 0.709 \pm 0.025 $  & - & $ 0.883 \pm 0.036 $  & - & $ 0.728 \pm 0.031 $ & - \\
        \bottomrule
    \end{tabular}
    }
\end{table}

The above analysis reveals the superiority of deep learning methods and GNNs, advancing the predictions of particle crushing strength by a large margin.
The ablation study aims to investigate the effects of particle morphology descriptors (PMD) and node and edge features (NEF) for GNNs.
We further compute the largest eight distances of a particle fragment to other fragments as the features of node distance and extract the overall largest eight distances as the particle features to complement the PMD and NEF.
In this way, \textit{w/o PMD} means removing the PMD or the PMD and the particle distances at the same time, and \textit{w/o NEF} means removing the NEF from GNNs, leaving the features of node distance for GNNs.
In summary, PMD has dominated the prediction accuracy of GNNs due to the expert experience~\cite{blott2008particle,bagheri2015characterization,huang2020experimental,domokos2015universality,pouranian2020impact,zhao20163d,yang2017effects,zheng2021three,zhang2016preliminary}, and NEF can further substantially reduce the prediction errors of GNNs consistently on various GNN architectures and the prediction tasks.

Since GNNs exhibit larger variances under different random seeds as shown in~\reftab{tab:main-result}, we visualize the variance ranges of GNNs under different parameter settings in~\reffig{fig:ablation} to evaluate the global effects of the PMD and NEF on the hyper-parameter sensitivity of different GNNs on different tasks.
Firstly, \reffig{fig:ablation} shows the significance of PMD that the prediction errors across three tasks increase rapidly with GNNs \textit{w/o PMD}, and the superiority of ExpC that ExpC mostly achieves the best medium performance of the boxplot against two other methods due to its powerful expressivity~\cite{yang2020breaking}.
Secondly, PMD reduces the variances of different parameter settings much more than NEF.
Thirdly, GNNs equipped with NEF obtain lower MAE than GNNs \textit{w/o NEF} as shown in the MAE distribution, especially on the Shape and Axis tasks of ExpC.
It implies that we should identify the task difficulty first and then choose the appropriate GNN architecture.

The best MAE comparison of ablation studies in~\reftab{tab:ablation} obtains consistent results with~\reffig{fig:ablation} that both PMD and NEF reduce the prediction errors of GNNs across the three tasks, and PMD has a much more significant impact than NEF on the three tasks, which are quantitatively verified by the degradation percentages.
In contrast to the similar performance of GNNs \textit{w/o NEF} and Baseline in~\reffig{fig:ablation}, \reftab{tab:ablation} provides a more detailed observation that Baseline can improve GNNs \textit{w/o NEF} by up to 13.2\% and 12.0\% on the Shape and the Axis task, indicating the effectiveness of NEF.
Moreover, the best-performing ExpC benefits from NEF consistently across the three tasks.
Overall, the ablation studies verified the instability of GNNs without the dedicated PMD and the effectiveness of NEF when applied to GNNs.

\subsection{Attribution Analysis}



\begin{figure}[htbp]
    \centering
    \includegraphics[width=\linewidth]{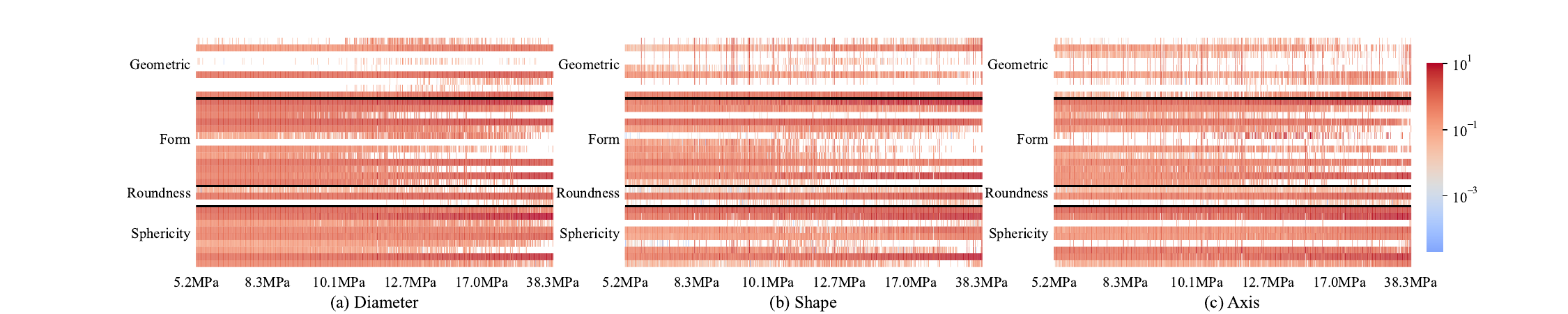}
    \caption{Gradient attribution of the first 34 particle morphology descriptors.}
    \label{fig:graph-grad}
\end{figure}

\begin{figure}[htbp]
    \centering
    \includegraphics[width=\linewidth]{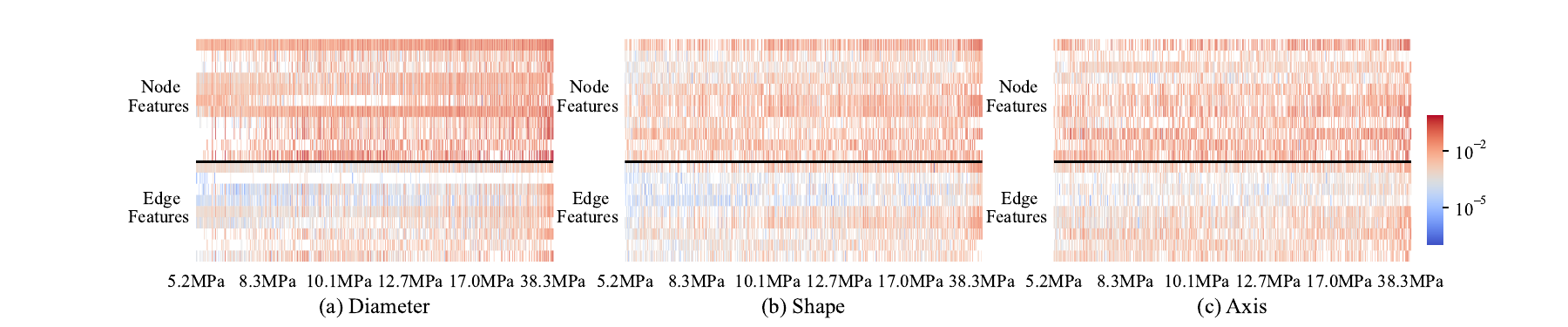}
    \caption{Gradient attribution of node and edge features.}
    \label{fig:node-grad}
\end{figure}

The ablation studies reveal the significance and sensitivity of particle morphology descriptors (PMD) and node and edge features (NEF) with respect to different tasks and GNNs.
The attribution analysis aims to understand the detailed mechanisms of PMD and NEF for particle crushing strength prediction with the common patterns and individual patterns of the feature attributions on the three tasks.
We adopt the simple yet effective gradient-based method to quantify the impacts of inputs for the predictions, which has been employed from CNNs~\cite{song2019deep,feng2022model} to GNNs~\cite{jing2021amalgamating}.
We choose the ExpC with its best-performing hyper-parameters as shown in~\reftab{tab:main-result} and fix its random seed as 0 to obtain the trained model.
Let $\hat{\rvy}$ be the prediction strength of a particle and $\rvx$ be the input PMD and NEF, and we average the magnitudes of gradient attributions for a specific particle type, written as $\frac{1}{\vert \bfX_{type} \vert} \sum_{\rvx \in \bfX_{type}} \lvert \frac{\partial \hat{\rvy}}{\partial \rvx} \rvert$, where $\bfX_{type}$ is a particle type with the specific diameter, shape, and axis.
The attribution distributions of PMD and NEF are visualized in~\reffig{fig:graph-grad} and~\reffig{fig:node-grad}, where the X-axis represents the true particle strength from 5.09 MPa to 38.29 MPa, including the training set, the validation set and the testing set to eliminate the differences of the testing sets of the three tasks and discover the common and individual patterns of the attribution distributions.

We remove the 35-th PMD of the stability descriptor since its small attributions degrade the visualization effects, which is provided in~\reffig{fig:graph-grad-full}.
The first 34 PMD are divided into four groups as shown in~\reffig{fig:graph-grad}, \textit{e.g.} Geometric, Form, Roundness, and Sphericity.
The highlighted horizontal bars shown in~\reffig{fig:graph-grad} summarize the common patterns of the three tasks that only partial features of different groups provide impressive attributions.
On the one hand, it is caused by the similarity and redundancy of different descriptors.
On the other hand, the simulation particles exhibit the standard spheroidicity with specified shape factors rather than realistic particles with various shapes~\cite{22wang2021machine,ma2022morphology}, which limits the effectiveness of morphology descriptors.
Besides the common patterns, the sparse vertical bars occur more frequently on the Shape and the Axis task than on the Diameter task, which is consistent with the task difficulty as shown in~\reftab{tab:main-result} that a difficult task requires more complex relationships between different descriptors.
The horizontal and vertical bars can point out the effectiveness and the cooperative relationships of PMD on the three tasks.

NEF contain the local descriptors of particle fragments and their interactive faces.
As depicted in~\reffig{fig:node-grad}, the attribution distributions of NEF exhibit significantly smaller values than those of PMD.
This difference can be attributed to the hierarchical aggregation mechanism of ExpC, which aggregates the small attributions to large attributions, and the higher significance of PMD in the hybrid framework, as shown in~\reftab{tab:ablation}.
Although the node features provide more attributions than the edge features, the node features exhibit larger attributions from the Diameter task to the Shape and Axis task, while the edge features exhibit the contrary trend.
It implies that the interactions of fragments can well model the particle crushing behaviors with the \sota~GNNs on the more difficult tasks.
It will be a promising direction to explore the interactions of fragments during particle crushing in future work.

\section{Discussion}

Existing researches have developed various statistical theories like Weibull statistics $P_s(V_0) = \exp \left[ - \left( \frac{\sigma}{\sigma_0} \right)^m \right]$ and machine learning methods to characterize the size effects~\cite{sun2021influence,hu2022particle} and morphology effects~\cite{22wang2021machine,ma2022morphology} of particle crushing.
However, since particle crushing occurs in various civil engineering fields~\cite{xiao2020grain}, GNNs with the fragment graph, as constitutive modeling, can largely alleviate the efforts to predict the particle crushing strength more accurately and attribute the prediction contributions more efficiently.
We summarize the limitations and future works as follows.

\textit{More realistic simulations.}
We generate the particle crushing dataset following Hu~\etal~\cite{hu2022particle}, which varies the diameter, shape, and axis as shown in~\reftab{tab:dataset}.
Nonetheless, the dataset mainly consists of particles with the standard spheroidicity, which deviate largely from the realistic particle shapes~\cite{sun2021influence,22wang2021machine,ma2022morphology}.
Besides, the uniaxial compression is employed to simplify the loading condition in the particle crushing process, while the triaxial test~\cite{hu2022particle} and different loading conditions can be explored in the future. 

\textit{More powerful GNNs for particle crushing.}
We test the generalization ability of these \sota~GNNs by reducing the learning difficulty of GNNs based on the dedicated PMD features~\cite{22wang2021machine}, which leaves space for improvements.
For example, the interactions of fragments during loading can be modeled by GNNs to explore the detailed behaviors of particle crushing in different time steps.  
Besides, it is necessary to consider the external loading conditions when introducing more realistic simulations.

\textit{Quantitative attributions of features.}
We explain the feature attributions based on the quantitative gradient-based methods~\cite{song2019deep,jing2021amalgamating,feng2022model}.
However, the common and individual patterns of different tasks are qualitatively summarized to highlight the contributions of different features.
It requires cooperation with civil engineering researchers to iteratively identify the significant descriptors and predict the particle crushing strength with GNNs.

\section{Conclusion}

In this paper, we are motivated by the lack of a large-scale dataset on particle crushing and the successful applications of GNNs to propose a method to predict particle crushing strength based on fragment connectivity.
Experimental results demonstrate the effectiveness of a hybrid framework that combines MLP and GNNs, and surprisingly show the comparable performance of three different GNNs, namely MeshNet~\cite{feng2019meshnet}, GIN~\cite{xu2018how}, and ExpC~\cite{yang2020breaking}.
The significantly improved performance motivates further exploration of particle crushing under different conditions, such as variations in particle diameters, shapes, and compression axes. 
Our analysis reveals the importance and sensitivity of particle morphology descriptors and features of GNNs in generalizing to different tasks.
Nonetheless, in everyday use of granular materials, various compression conditions require more comprehensive modeling of particle crushing.
In the future, we aim to explore the use of Graph Neural Networks (GNNs) with improved connectivity to predict particle crushing based on a more comprehensive modeling approach.

\appendix

\section{Particle Crushing Dataset}

\begin{figure}[t]
    \centering
    \includegraphics[width=.6\textwidth]{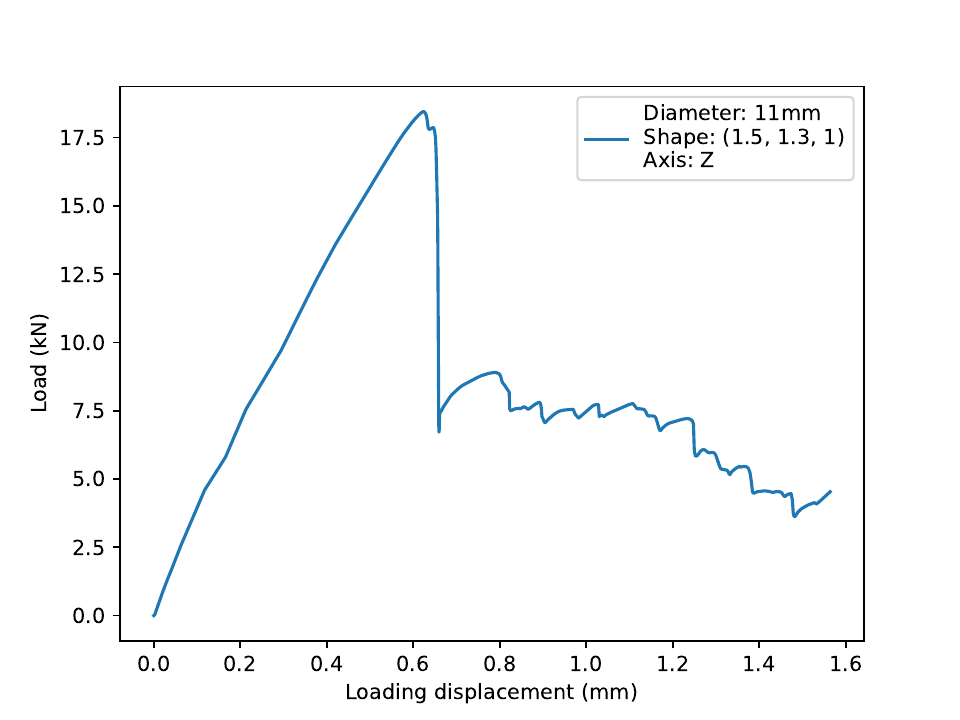}
    \caption{The load-displacement curve in a particle compression test.}
    \label{fig:sample-curve}
\end{figure}

Figure~\ref{fig:sample-curve} describes the load-displacement curve of a randomly sampled particle, where the peak force decreases drastically due to the particle breakage during the external load.
We strictly follow the instructions of Hu~\etal~\cite{hu2022particle} to perform a batch of simulations of 45,000 particles, adopting the Non-Smooth Contact Dynamics (NSCD) method for the numerical simulations of particle crushing.
The parameters of the Cohesive Zone Model (CZM) of the NSCD method are also calibrated by Hu~\etal~\cite{hu2022particle}.
After generating the comprehensive dataset, we filter particles without at least 35\% decrease~\cite{22wang2021machine,26huillca2021modelling,hu2022particle} from the peak force since these particles aren't considered to break into pieces, and compute the characteristic particle crushing strength $\sigma_0$ of 900 particle types, where the particle types without more than 30 valid simulations~\cite{mcdowell2000application,hu2022particle} of particle crushing are further removed to ensure the data sufficiency for the fitted Weibull distribution.
Finally, we have obtained 40,818 valid data points and 882 particle types of particle crushing.

\subsection{Compression Test of Single Particle}

\textit{Particle meshing.}
The software Neper~\cite{29quey2011large} provides the Voronoi tessellation of a sphericity particle, whose diameter, three-dimensional scale coefficients, and rotation angles are given as the input parameters of Neper.
The different diameters, shapes, and compression axes of our dataset are firstly introduced in the particle meshing process.
The Voronoi tessellation divides the specified polyhedral particle into multiple convex cells, simulating the fragment distribution of the particle and the common faces of different fragments.
A Voronoi cell can be defined as follows~\cite{cantor2017three}
\begin{equation}
    V_i = \{x \in \bfX \,\vert\, d(x, P_i) < d(x, P_j) \, \forall j \neq i\},
\end{equation}
where $\bfX$ is the three-dimensional point set of the particle space, $P_i$ and $P_j$ are the given points of the $i$-th and $j$-th cell.
The final output of Voronoi tessellation is optimized by iteratively replacing the given points $P_i, P_j$ with their new centroids $c_i, c_j$ until convergence~\cite{cantor2017three}.
Gmsh~\cite{gmsh} further produces a meshing of the particle based on the \textbf{geo} file generated by Neper.
We have also obtained the \textbf{tess} file generated by Neper for the convenience of modelling the fragments and their contacts in a fragment graph.

\begin{table}
    \centering
    \caption{Parameters for the cohesive zone model.}
    \label{tab:czm} 
    \resizebox{\textwidth}{!}{
    \begin{tabular}{cccccccc}
        \toprule
        $K_I \mathrm{(N/mm^3)}$ & $K_{II} \mathrm{N/mm^3}$ & $\sigma_I^e \mathrm{MPa}$ & $\sigma_{II}^e \mathrm{MPa}$ 
        & $G_I \mathrm{mJ/mm^2}$ & $G_{II} \mathrm{mJ/mm^2}$ & $\mu$ & $\rho \mathrm{kg/m^3}$ \\
        \hline 
        80 & 120 & 9 & 11.5 & 900 & 1125 & 0.3 & 2650 \\
        \bottomrule
    \end{tabular}
    }
\end{table}

\textit{Numerical simulation.}
We perform numerical simulations of the NSCD method by LMGC90~\cite{dubois2013lmgc90} after generating the mesh files of 45,000 particles.
The meshing particle with a specified rotation axis is assumed to be compressed along the new Z-axis by two parallel platens.
The NSCD method~\cite{31jean1999non,jean1992unilaterality,moreau1988unilateral} updates the displacement of each element at each time step by detecting the contacts and computing the contact forces between bodies, following the Signorini-Coulomb condition for a unilateral contact.
The unilateral Signorini condition describes the non-penetrability between two objects, written as follows
\begin{equation}
    \begin{cases}
        v_n > 0 &\Rightarrow f_n = 0, \\
        v_n = 0 &\Rightarrow f_n >= 0,
    \end{cases}
\end{equation}
where $v_n$ is the relative velocity between particles in the normal direction, and $f_n$ is the contact force in the normal direction.
The contact force $f_t$ in the tangential direction obeys the Coulomb friction law, written as
\begin{equation}
    \begin{cases}
        v_t > 0 &\Rightarrow f_t = -\mu f_n, \\
        v_t = 0 &\Rightarrow -\mu f_n \le f_t \le \mu f_n, \\
        v_t < 0 &\Rightarrow f_t = \mu f_n,
    \end{cases}
\end{equation}
where $\mu$ is the contact friction coefficient between particles and is set as 0.3 in our simulations.
The CZM model~\cite{camacho1996computational,hu2022particle} quantifies the relationship between breakage energy and the maximum displacement of the breakage of the contact faces, whose parameters are determined based on the work of Hu~\etal~\cite{hu2022particle}. 

\subsection{Weibull Distribution For Particle Crushing}

\begin{figure}[t]
    \centering
    \includegraphics[width=.6\textwidth]{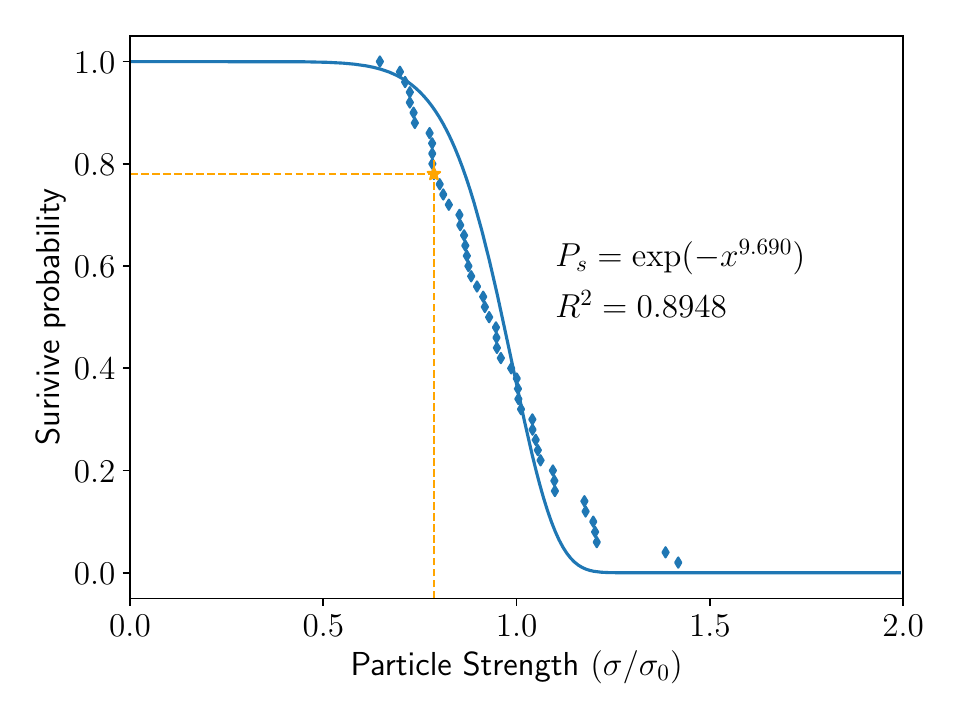}
    \caption{50 particle crushing data points and the fitting Weibull distribution of crushing strength. The particle type and the highlighted data point are both the same with \reffig{fig:sample-curve}.}
    \label{fig:sample-weibull}
\end{figure}

McDowell~\etal~\cite{mcdowell2000application} have found that the particle crushing strength of geometrically similar particles follows a Weibull distribution, written as
\begin{equation}
    P_s = \exp \left[ - \left(\frac{d_s}{d_0}\right)^3 \left(\frac{\sigma}{\sigma_0}\right)^m \right],
    \label{eq:weibull}
\end{equation}
where $P_s$ is the breakage ratios under the corresponding load, $d_0$ and $\sigma_0$ are the characteristic diameter and crushing strength, respectively.
The survival probability of a batch of 50 particles with the same geometry is drawn in \reffig{fig:sample-weibull}, sorting by the normalized particle strength, where $m=9.690$ is the Weibull modulus and indicates the variability of the particle crushing distribution.
The characteristic particle strength is thus scaled to 1.0 of the horizontal axis, and corresponds to the 37\% survival probability of the vertical axis.
Existing particle crushing theories~\cite{mcdowell2000application,sun2021influence,22wang2021machine} derive a close-form formula to describe the size effects of particles with different diameters.
However, firstly, their various theories have been proposed from different perspectives and lacked the consensus of the particle crushing behaviors.
Secondly, our experimental results of the non-Deep methods and DNNs in \reftab{tab:main-result} reveal that particle morphology descriptors play a vital role on the generalization of predicting the particle crushing strength, while existing theories are mostly based on the particle diameter.
Thirdly, generalizing the predictions for more conditions like the particle shapes and compression axes require a constitutive model to describe the particle from different viewpoints.
Therefore, we are motivated to model the fragment connection of the particle with Graph Neural Networks (GNNs), which offer a natural modelling for the fragment graph.

\subsection{Dataset Statistics}

\begin{figure}[t]
    \centering
    \includegraphics[width=.6\textwidth]{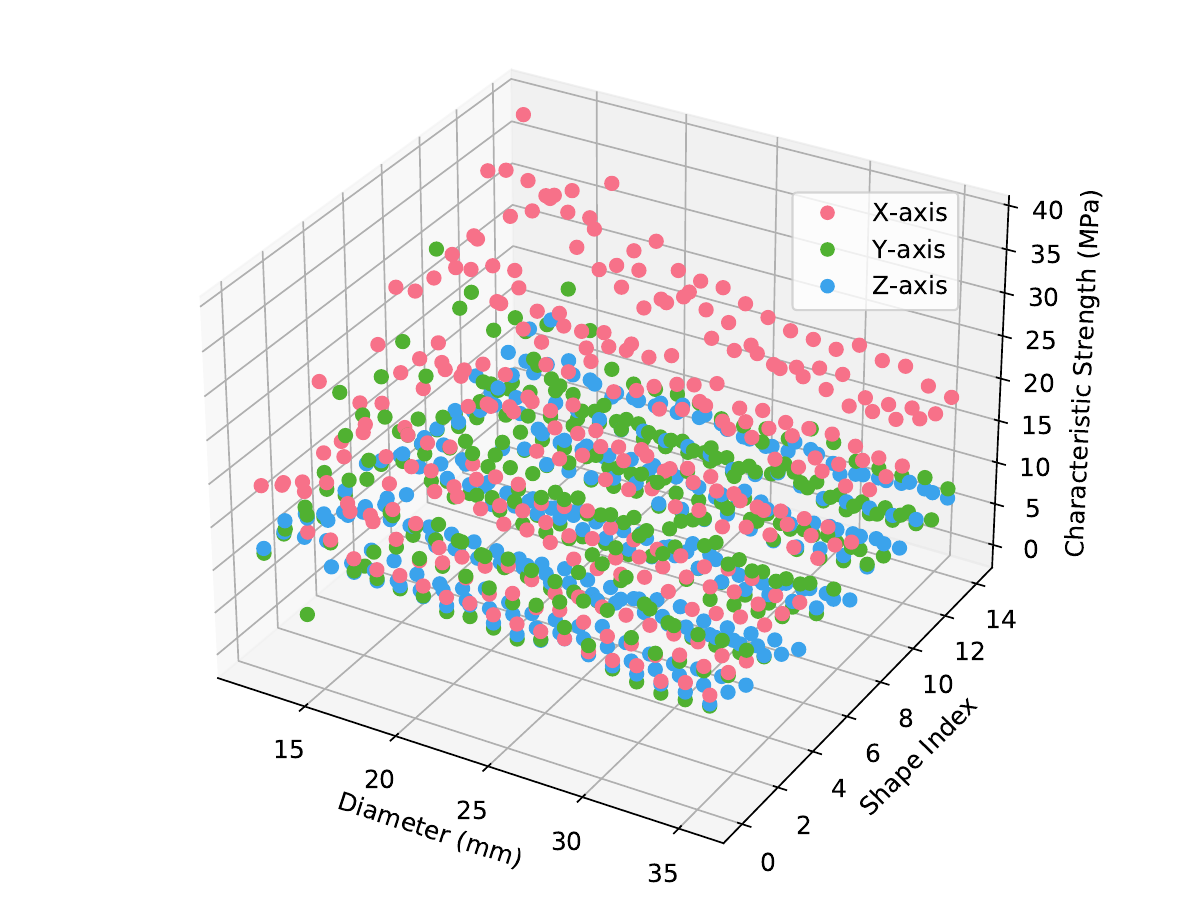}
    \caption{The characteristic particle crushing strength with respect to different diameters, shapes and compression axes.}
    \label{fig:charac3D}
\end{figure}

\begin{table}[t]
    \caption{The distribution of the characteristic strength ($\sigma_0$) and Weibull modulus $m$ of the dataset.}
    \centering
    \label{tab:data-stat} 
    \begin{tabular}{lccccc}
        \hline
         & min & 25\% & 50\% & 75\% & max \\
        \hline
        $\sigma_0$ (MPa) & 5.09 & 8.65 & 11.03 & 15.06 & 38.29 \\
        $m$ & 0.69 & 3.88 & 5.54 & 6.86 & 11.60 \\
        \hline
    \end{tabular}
\end{table}

In the particle generation stage, we have chosen 900 particle types, which is the Cartesian product of 20 particle diameters, 15 scale shapes in the (X, Y, Z) axes, and 3 compression axes, as shown in \reftab{tab:dataset}.
We have conducted 50 numerical simulations with random Voronoi tessellation for each particle type to ensure the data sufficiency to measure the characteristic crushing strength.
Firstly, we reserve the particles that can be observed at least 35\% decrease~\cite{22wang2021machine,26huillca2021modelling,hu2022particle} from the peak force to ensure the particle breakage.
Secondly, we remove particle types that contain less than 30 valid simulations~\cite{mcdowell2000application,hu2022particle} from our dataset to ensure the data quality.
As stated before, we have obtained 882 particle types and 40,818 particles finally.

\reftab{tab:data-stat} describes that 882 particle types present smooth variations in terms of the characteristic strength $\sigma_0$ and the Weibull modulus $m$, which is consistent with observations from~\cite{sun2021influence,hu2022particle}.
Moreover, the outliers of the particle types are centered around the maximum values, where the maximum values are around two times larger than the 75\% percentile of $\sigma_0$ and $m$.
It poses significant challenges to the machine learning methods.
\reffig{fig:charac3D} visualizes the distribution of $\sigma_0$ in terms of three different conditions, including the diameter, shape and compression axis.
Firstly, the compression axis increases $\sigma_0$ more significantly than the other two conditions, which may be attributed to that the scale coefficient of the X-axis is larger than the Y-axis and Z-axis as shown in \reftab{tab:dataset}.
Secondly, the outliers of the 882 particle types are centered in the corner of the 3D plot, which presents small diameter, flat shape and the X-axis.
We believe that the comprehensive dataset can sufficiently reveal the diversity of particle crushing behaviors and benchmark the machine learning methods for particle crushing.

\subsection{Particle Features}

\begin{figure}[h]
    \centering
    \includegraphics[width=.9\textwidth]{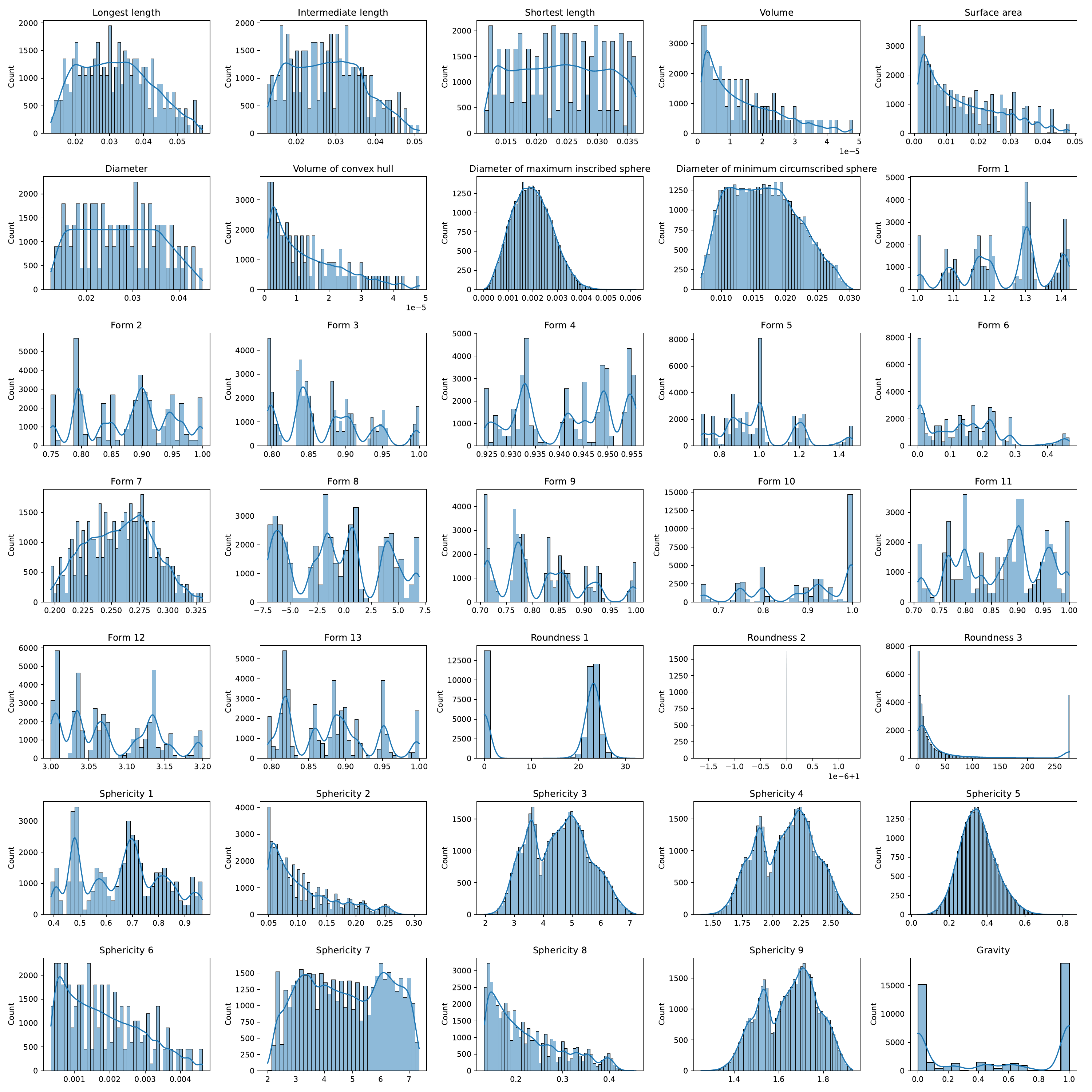}
    \caption{The distribution of 35 Particle Morphology Descriptors (PMD)~\cite{22wang2021machine}, including 9 geometric characteristics~\cite{22wang2021machine}, 13 form descritpors~\cite{blott2008particle,bagheri2015characterization,huang2020experimental,domokos2015universality}, 3 roundness descritpors~\cite{pouranian2020impact,huang2020experimental,zhao20163d}, 9 sphericity descriptors~\cite{blott2008particle,pouranian2020impact,yang2017effects,zheng2021three,huang2020experimental,zhang2016preliminary}, and 1 instability descriptor~\cite{22wang2021machine}.}
    \label{fig:features}
\end{figure}

We have calculated thirty-five Particle Morphology Descriptors (PMD) according to the work of Wang~\etal~\cite{22wang2021machine}, as shown in \reffig{fig:features}.
The feature distributions can be roughly categorized into three kinds: the uniform distribution (\textit{e.g.}, Shortest Length), the long-tail distribution (\textit{e.g.}, Volume), and the Gaussian distribution (\textit{e.g.}, Diameter of the maximum inscribed sphere).
It is noteworthy that the distributions of the flatness (Form 11) and the sphericity (Sphericity 8) are significantly different from the Gaussian distribution of that~\cite{22wang2021machine}, which is caused between the differences of simulated particles and natural particles.
The diverse feature distributions also verify that it is rather difficult to predict particle crushing strength only based on the diameter feature. 

\begin{table}[t]
    \caption{Feature components of node features and edge features, where (3) means three variables.}
    \label{tab:features}
    \centering
    \begin{tabular}{cp{0.8\linewidth}}
        \toprule
        Features & Details \\
        \hline
        Node Features & Volume(1), Surface Area(1), Diameter(1), Number of Faces(1), Number of Neighbors(1), Centroid Coordinates(3), Euler Angles(3) \\
        \hline
        Edge Features & Contact Area(1), Number of Lines(1), Maximum Length of the Contact Area(1), Centroid Coordinates(3), Euler Angles(3) \\
        \bottomrule
    \end{tabular}
\end{table}

We have also developed eleven node features to describe each fragment of the particle and nine edge features for the connectivity between fragments, as shown in \reftab{tab:features}.
These features also present similar distributions to the PMD since they share similar definitions with PMD.
The particular advantage of node features and edge features is that they can describe the microscopic behaviors of fragments beyond the overall behaviors of the particle.

\section{Graph Neural Networks for Particle Crushing}

\subsection{Graph Neural Networks}

Graph Neural Networks (GNNs) have shown its powerful prediction abilities on the non-grid data and been applied to various fields such as Recommendation~\cite{ying2018graph}, traffic flow prediction~\cite{yu2018spatio}, and molecular classification~\cite{gilmer2017neural,xu2018how,yang2020breaking,ying2021transformers}.
A graph is defined by a node set $V=\{v_1, \cdots, v_n\}$ and an associated edge set $E=\{e_1, \cdots, e_m\}$, where $e_1$ connects a pair of adjacent fragments $v_i$ and $v_j$.
The message-passing paradigm~\cite{gilmer2017neural} describes GNNs in a microscopic view that each node $v_i$ receives the messages from its neighbors $v_j \in \mathcal{N}(v_i)$ and updates its representation based on these messages, written as:
\begin{equation}
    \begin{split}
       \rvm_i^k &= \sum_{v_j \in \mathcal{N}(v_i)} \bfM_k(\rvh_i^k, \rvh_j^k, \rve_{ij}), \\
       \rvh_i^{k+1} &= \bfU_k(h_i^k, \rvm_i^k),
    \end{split}
\end{equation}
where $v_i$ updates from $\rvh_i^{k}$ of the $k$-th layer to $\rvh_i^{k+1}$ of $k+1$-th layer based on the message function $\bfM_k$ and the update function $\bfU_k$.
The obtained node representation $\rvh_i$ can be used for various tasks such as node classification~\cite{kipf2016semi} and link prediction~\cite{ying2018graph}.
It requires a readout function to generate a representation vector for the whole graph $G$ for graph-level tasks like molecular classification~\cite{gilmer2017neural,xu2018how,yang2020breaking,ying2021transformers}, written as
\begin{equation}
    \rvh_G = \bfR(\{\rvh_i^K \vert v_i \in V\}),
\end{equation}
where $K$ is the number of neural network layers. 

We have adopted MeshNet~\cite{feng2019meshnet}, GIN~\cite{xu2018how}, and ExpC~\cite{yang2020breaking} as our baseline GNNs for particle crushing.
We have also tried the powerful Graphormer~\cite{ying2021transformers}, which has achieved the first place together with ExpC~\cite{yang2020breaking} in the PCQM4M-LSC track of KDD Cup 2021~\footnote{https://ogb.stanford.edu/kddcup2021/results/\#awardees\_pcqm4m}
, but failed the memory requirements on our GPU of 24GB Titan X.
These three GNNs can be summarized as follows:
\begin{itemize}
    \item MeshNet~\cite{feng2019meshnet} is called a Convolutional Neural Network for meshing files, by aggregating neighborhood information based on its spatial descriptor and structural descriptor and pooling the overall representation. 
    The working mechanism of MeshNet~\cite{feng2019meshnet} exactly meets the definition of the message-passing paradigm~\cite{gilmer2017neural} and implements the readout function with a global average pooling layer.
    Therefore, MeshNet is considered as a variant of GNNs in this paper.
    \item GIN~\cite{xu2018how} is proposed as a more powerful GNNs to discriminate the isomorphism graph better by building an injective mapping from neighborhood message to node representations, innovatively leading a new direction of enhancing the discriminative power of GNNs.
    \item ExpC~\cite{yang2020breaking} improvess the expressive power of GNNs beyond 1-WL test based on the innovative ExpandingConv and CombConv.
    Ying~\etal~\cite{ying2021transformers} have achieved the first place in the PCQM4M-LSC track of KDD Cup 2021 by combing Graphormer~\cite{ying2021transformers} and ExpC~\cite{yang2020breaking}.
\end{itemize}

\subsection{Predicting Particle Crushing Strength}

The crushing behavior of a particle is closely related with the particle shape and the internal structure, as shown in previous theoretical studies~\cite{2mcdowell1998micromechanics,7daouadji2001elastoplastic,8russell2004bounding,9tengattini2016constitutive}.
Numerical simulations of various particles have become the mostly adopted method to study the relations between the particle and its crushing strength since experimental tests are costly and unscalable~\cite{13cheng2004crushing,14de2014triaxial,15fu2017discrete}.
In a numerical simulation, a particle is divided into multiple rigid polyhedral cells randomly~\cite{28nguyen2015bonded} and gets fractured during the compression test~\cite{hu2022particle}.
The crushing strength of the particle reveals its internal characteristic, benefiting for various engineering fields including geotechnical engineering, mining, chemical engineering, and so on~\cite{3zhu2019modeling}.

From the perspective of machine learning, the tessellated particle can be seen as a connected graph, where each polyhedral cell plays as a node.
Formally, the cell set is denoted by $V=\{v_1, \cdots, v_i, \cdots, v_n\}$, where $v_i$ refers to a polyhedral cell and $n$ denotes the number of cells.
A node $v_i$ is usually associated with a node vector $\rvv_i$ to describe the node states, as listed in \reftab{tab:features}.
The edges of the \emph{internal graph} are formed according to the spatial proximity of each cell pair.
The edge set of the internal graph is denoted by $E=\{e_1, \cdots, e_m\}$, where $e_1$ connects two in-contact (adjacent) cells $v_i$ and $v_j$.
An edge $e_k$ could also carry the relational vector $\rve_k$ of the two cells, such as the contact surface areas and the angle of the contact~\cite{42vlassis2020geometric}.
Our task is predicting the particle crushing strength on particles of unseen types with the help of GNNs~\cite{feng2019meshnet,xu2018how,yang2020breaking}, given the internal graph $G=\{V, E\}$ and its associated vectors $\rvx = \{\{\rvv_i \vert \forall v_i \in V\}, \{\rve_k \vert \forall e_k \ in E\}\}$.
The problem can be formulated as 
\begin{equation}
    \argmin_{\rvg}  \sum_{\rvx \in \bfX, \rvy \in \bfY} \lrVert{\rvy - \rvg(\rvx, G)}_2 + \lambda\Omega(\rvg),
\end{equation}
where $\rvg$ is one kind of GNNs and $\Omega$ is the regularization term.


\section{Experiment Settings}

We have obtained three testing sets by splitting the dataset according to the diameters, scale shapes, and compression axes to examine the generalization ability of different methods across three tasks.
The testing sets are split around 1/3 as shown in \reftab{tab:dataset}, where 7/20 for the diameter task, 5/15 for the shape task, and 1/3 for the axis task.

\subsection{Hyper-parameters}

\begin{table}
    \centering
    \caption{Grid search of parameters of machine learning methods.}
    \label{tab:nondeep-models} 
    \resizebox{\textwidth}{!}{
    \begin{tabular}{cll}
        \toprule
        Method & Parameter & Settings \\
        \hline
        Linear & Default & Default \\
        \hline
        Ridge & alpha & [1e-3, 1e-2, 1e-1, 1.0] \\
        \hline
        \multirow{3}{*}{Random Forest} & n\_estimators & [10, 25, 50, 75, 100] \\
        & min\_samples\_split & [2, 8, 32] \\
        & criterion & [squared\_error, absolute\_error, poisson] \\
        \hline
        \multirow{4}{*}{LightGBM} & n\_estimators & [10, 25, 50, 75, 100] \\
        & max\_depth & [4, 16, 64] \\
        & reg\_alpha & [0.0, 1e-3, 1e-2, 1e-1, 1.0] \\
        & reg\_lambda & [0.0, 1e-3, 1e-2, 1e-1, 1.0] \\
        \hline
        \multirow{4}{*}{XGBoost} & n\_estimators & [10, 25, 50, 75, 100] \\
        & max\_depth & [4, 16, 64] \\
        & reg\_alpha & [0.0, 1e-3, 1e-2, 1e-1, 1.0] \\
        & reg\_lambda & [0.0, 1e-3, 1e-2, 1e-1, 1.0] \\
        \bottomrule
    \end{tabular}
    }
\end{table}

\begin{table}
    \centering
    \caption{Default parameters of deep learning methods.}
    \label{tab:default-deep-models} 
    \resizebox{\textwidth}{!}{
    \begin{tabular}{cp{0.9\linewidth}}
        \toprule
        Method & Default Parameters \\
        \hline
        MLP & batch\_size=128, learning\_rate=1e-3, hidden=128, layers=2, dropout=1e-1 \\
        \hline
        MeshNet & batch\_size=128, learning\_rate=1e-3, hidden=128, layers=2, dropout=1e-1 \\
        \hline
        GIN & batch\_size=128, learning\_rate=1e-3, hidden=128, layers=2, dropout=1e-1, eps=1e-5, JK=max \\
        \hline
        ExpC & batch\_size=128, learning\_rate=1e-3, hidden=128, layers=2, dropout=1e-1, head\_size=32 \\
        \bottomrule
    \end{tabular}
    }
\end{table}

\begin{table}
    \centering
    \caption{Iterative search of parameters of deep learning methods.}
    \label{tab:deep-models} 
    \resizebox{\textwidth}{!}{
    \begin{tabular}{clp{0.5\linewidth}}
        \toprule
        Method & Parameter & Settings \\
        \hline
        \multirow{4}{*}{MLP} & batch\_size & [8, 32, 128, 512] \\
        & learning\_rate& [1e-5, 1e-4, 1e-3, 1e-2] \\
        & hidden & [128, 256, 512] \\
        & layers & [2, 3, 4, 5]\\
        \hline
        \multirow{4}{*}{MeshNet}& batch\_size & [8, 32, 128, 512] \\
        & learning\_rate& [1e-5, 1e-4, 1e-3, 1e-2] \\
        & hidden & [128, 256, 512] \\
        & layers & [2, 3, 4, 5]\\
        \hline
        \multirow{5}{*}{GIN} & batch\_size & [8, 32, 128, 512] \\
        & learning\_rate& [1e-5, 1e-4, 1e-3, 1e-2] \\
        & hidden & [128, 256, 512] \\
        & layers & [2, 3, 4, 5]\\
        & JK & [last, cat, max, lstm] \\
        \hline
        \multirow{4}{*}{ExpC} & batch\_size & [8, 32, 128, 512] \\
        & learning\_rate& [1e-5, 1e-4, 1e-3, 1e-2] \\
        & hidden & [128, 256, 512] \\
        & layers & [2]\\
        \bottomrule
    \end{tabular}
    }
\end{table}

We have conducted experiments with five non-Deep methods, including Linear Regression, Ridge Regression, Random Forest, LightGBM~\cite{lightgbm}, and XGBoost~\cite{xgboost}, and four DNNs, including MLP, MeshNet~\cite{feng2019meshnet}, GIN~\cite{xu2018how}, and ExpC~\cite{yang2020breaking}.
For a fair comparison, we have tried our best to grid search the hyper-parameters for non-Deep methods, as shown in \reftab{tab:nondeep-models}.
The experimental results of non-Deep methods verify the effectiveness of them based on the dedicated particle morphology descriptors.
Since the computation cost of DNNs is much larger than that of non-Deep methods, we firstly determine a group of parameters accroding to our experience as shown in \reftab{tab:default-deep-models}.
Particularly, the number of ExpC layers is set to 2 to avoid the Out-Of-Memory error on our GPU.
Then, we iteratively replace the default parameter with one parameter shown in \reftab{tab:deep-models}.
We have conducted five runs for each kind of parameter combination. 
In summary, we have conducted at least 15 runs for Linear Regression, 60 runs for Ridge Regression, 675 runs for Random Forest, 5625 runs for LightGBM, and 5625 runs for XGBoost across three tasks.
We have further performed three kinds of ablation studies for DNNs, resulting in 675 runs for MLP, 675 runs for MeshNet, 855 runs for GIN, and 540 runs for ExpC.
These experiments can largely reduce the variances of different methods and ensure a fair comparison.




\subsection{More results of Attribution Analysis}

\begin{figure}[htbp]
    \centering
    \includegraphics[width=\linewidth]{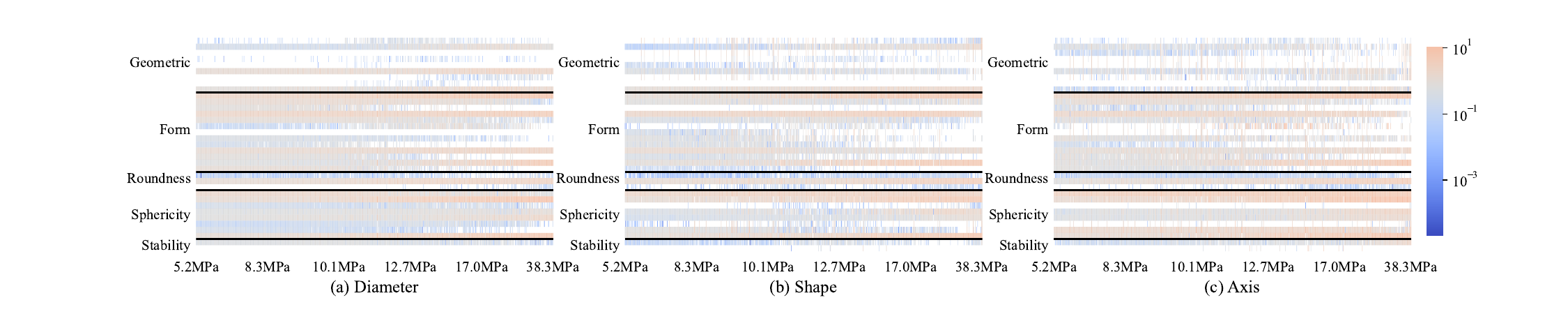}
    \caption{Gradient attribution of particle morphology descriptors.}
    \label{fig:graph-grad-full}
\end{figure}

As shown in \reffig{fig:graph-grad-full}, the 35-th PMD of the stability descriptor presents the smallest attributions with respect to different particle crushing strength, resulting in low contrast of different descriptors.
It can be observed from \reffig{fig:graph-grad-full} that the light and dark colors differ from each other significantly across the X-axis, which can be used for feature importance directly to remove less useful descriptors.
However, it poses challenges to compare the attribution analysis across the X-axis and three tasks.







\clearpage
\bibliography{ref}

\end{document}